%% file: main.tex
\documentclass[10pt,twocolumn,letterpaper]{article}

\usepackage[pagenumbers]{cvpr} 

\usepackage{graphicx}
\usepackage{amsmath}
\usepackage{amssymb}
\usepackage{booktabs}
\usepackage{caption}
\usepackage{subcaption}
\DeclareUnicodeCharacter{2212}{-}

\usepackage[pagebackref,breaklinks,colorlinks]{hyperref}

\usepackage[capitalize]{cleveref}
\crefname{section}{Sec.}{Secs.}
\Crefname{section}{Section}{Sections}
\Crefname{table}{Table}{Tables}
\crefname{table}{Tab.}{Tabs.}

\def\cvprPaperID{*****} 
\def\confName{CVPR}
\def\confYear{2023}

\begin{document}

\title{Emotion Recognition from the perspective
of Activity Recognition}

\author{Savinay Nagendra  \quad Prapti Panigrahi  \vspace{0.3em} \\
{\normalsize Department of Computer Science} \\
{\normalsize The Pennsylvania State University}\\
{\normalsize University Park}\\
{\tt\small\centering sxn265@psu.edu \quad pmp5425@psu.edu  \vspace{0.3em}}}
\maketitle

\begin{abstract}
   Automatic emotion recognition from using visual signals is one of the burgeoning research areas in the domain of Computer Vision. Applications of an efficient emotion recognition system can be found in several domains such as medicine, driver fatigue surveillance, social robotics and human-computer interaction. Appraising human emotional states, behaviors and reactions displayed in real world settings can be accomplished using latent continuous dimensions. Continuous dimensional models of human affect, such as those based on valence and arousal have been shown to be more accurate in describing a broad range of spontaneous everyday emotions than more traditional models of discrete stereotypical emotion categories (e.g. happiness, surprise). Most of the prior work on estimating valence and arousal consider laboratory settings and acted data. But, for emotion recognition systems to be deployed and integrated into real-world mobile and computing devices, we need to consider data collected in the wold. Action recognition is a domain of Computer Vision that involves capturing complementary information on appearance from still frames and motion between frames. In this paper, we treat emotion recognition from the perspective of action recognition by exploring the application of deep learning architectures specifically designed for action recognition, for continuous affect recognition. We propose a novel three stream end-to-end deep learning regression pipeline with attention mechanism, which is an ensemble design based on sub-modules of multiple state-of-the-art action recognition systems. The pipeline constitutes a novel data pre-processing approach with spatial self-attention mechanism to extract key frames. Optical flow of high attention regions of the face are extracted to capture temporal context. AFEW-VA in-the-wild dataset has been used to conduct comparative experiments. Quantitative analysis shows that the proposed model outperforms multiple standard baselines of both emotion recognition and action recognition models.
\end{abstract}

\input{intro}
\input{related}
\input{data}
\input{approach}
\input{results}
\input{conclusion}
\input{ack}
{\small
\bibliographystyle{ieee_fullname}
\bibliography{egbib}
}

\end{document}

%% file: intro.tex
\section{Introduction}\label{sec:intro}
The field of Automatic Emotion Recognition analysis has grown rapidly in recent years, with applications spread across a variety of fields, such as medicine \cite{41_tagaris2017assessment}, \cite{42_tagaris2018machine}, health monitoring  \cite{19_kollias2018deep}, entertainment, lie-detection \cite{50_zhou2015lie}, \cite{22_kollias2017adaptation}.  Current research in automatic analysis of facial affect aims at developing systems, such as robots and virtual humans, that will interact with humans in a naturalistic way under real-world settings. Such systems should automatically sense and interpret visual signals relevant to emotions, appraisals and intentions. There are two major emotion computing models according to theories in psychology research \cite{30_marsella2014computationally}: discrete and dimensional theory. When designing emotion recognition systems for real-world settings, where subjects operate in a diversity of contexts and environments, systems that perform automatic analysis of human behavior should be robust to the diversity of contexts,  the timing of display and video recording conditions.
\par There are two major emotion computing models according to theories
in psychology research \cite{30_marsella2014computationally}: discrete and dimensional theory. Research in the past \cite{nagendra2017comparison, nagendra2022constructing, funk2018learning,liu2021new,pei2021utilizing,nagendra2020cloud,nagendra2020efficient,nagendra2022threshnet,nagendra2024patchrefinenet,nagendra2023estimating,zhu2022rapid} has revolved around the recognition of the so-called six universal expressions - Happiness, Sadness, Fear, Disgust, Surprise and Anger. \cite{6_dalgleish2000handbook}, \cite{16_kollias2015interweaving}, \cite{25_kollias2018training}, \cite{3_cowie2003describing}, \cite{18_kollias2016line}. These are intuitive and simple, but cannot express complex affective states. For many years, research in automatic analysis of facial behavior was mainly limited to posed behavior which was captured in highly controlled recording conditions \cite{35_lucey2010extended, 41_tagaris2017assessment, 55_tian2001recognizing, 57_valstar2010induced}. Some representative datasets, which are still used in many recent works \cite{27_jung2015joint}, are Cohn-Kanade database \cite{35_lucey2010extended, 55_tian2001recognizing}, MMI database \cite{41_tagaris2017assessment, 57_valstar2010induced}, Multi-PIE database \cite{22_kollias2017adaptation} and the BU-3D and BU-4D databases \cite{62_zhang2013high, 63_yin20063d}. 
\par However, the fact that the facial expressions of naturalistic behaviors can be radically different from the posed ones has been widely accepted in the research community \cite{10_corneanu2016survey, 48_sariyanidi2014automatic, 66_zeng2008survey}. Hence, efforts
have been made in order to collect subjects displaying
naturalistic behavior. Examples include the recently collected
EmoPain \cite{4_aung2015automatic} and UNBC-McMaster \cite{36_lucey2011painful} databases for
analysis of pain, the RU-FACS database of subjects participating
in a false opinion scenario \cite{5_bartlett2006fully} and the SEMAINE corpus
\cite{39_mckeown2011semaine} which contains recordings of subjects interacting
with a Sensitive Artificial Listener (SAL) in controlled conditions.
All the above databases have been captured in well-controlled
recording conditions and mainly under a strictly
defined scenario eliciting pain.
\begin{figure}[!t]
    \centering
    \includegraphics[width=3in]{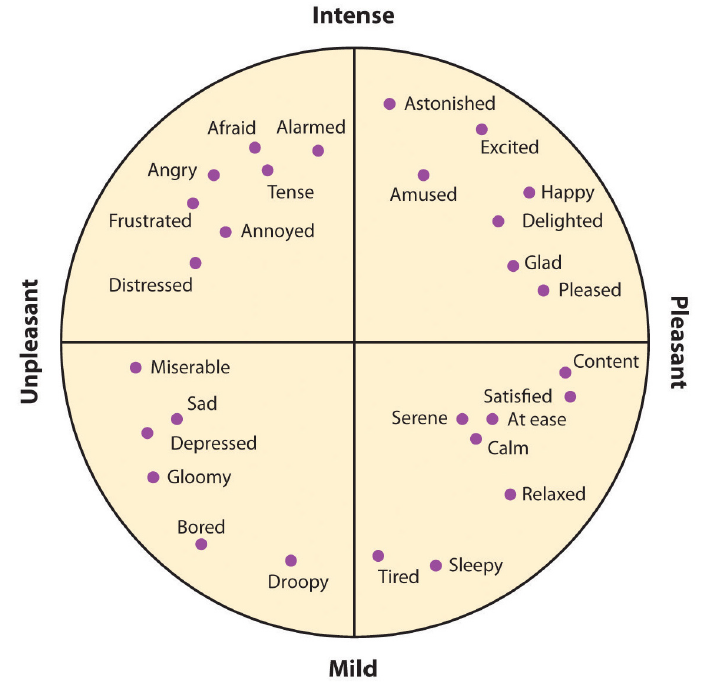}
    \caption{The 2D Emotion wheel or the 2D Valence-Arousal Space}
    \label{fig:wheel}
\end{figure}
\par Valence defines how negative or positive the experience is, and intensity of arousal defines how calming or exciting the experience is. The circumplex model of emotion shown in Figure \ref{fig:wheel} represents the 2D Emotion Wheel or the 2D Valence-Arousal Space. The dimensions represent arousal, indicating the
level of affective activation, and valence, measuring the level
of pleasure. Hence, dimensional theory can model subtle,
complicated, and continuous affective behaviors.
\par Action recognition task involves the identification of different actions from video clips (a sequence of 2D frames) where the action may or may not be performed throughout the entire duration of the video. This seems like a natural extension of image classification tasks to multiple frames and then aggregating the predictions from each frame. Despite the stratospheric success of deep learning architectures in image classification (ImageNet), progress in architectures for video classification and representation learning has been slower. Both Emotion and Action recognition algorithms are similar in terms of:(1)
\textbf{Input}: (a)Input for both tasks is a sequence of 2D video frames. (b) The action/emotion may or may not be performed throughout the entire video. (c) Multiple actions/emotions possible in the same video. (2) \textbf{Capturing Long Context}: (a) Have to capture spatio-temporal context across frames. (b) Spatial information captured has to be compensated for camera movement. (c) Both local and global context with respect to motion information needs to be captured for robust predictions. (3) \textbf{Architectures and Training}: (a)One network for capturing spatio-temporal information. (b) Two separate networks, for each spatial and temporal. (c) Fusing predictions across multiple frames. (d) End-to-end training vs feature extraction and classifying separately. 
\par The motivation to explore this problem comes from the fact that the field of emotion recognition is nascent and not as advanced as the field of action recognition in the domain of Computer Vision. With this research we seek the answers to the following questions: (1) With the above mentioned similarities between the two tasks, is it feasible to explore emotion recognition from the perspective of Action Recognition? (2) Can the complex deep learning mechanisms of Action recognition be adopted for the task of emotion recognition? (3) Can we come up with a general model which can work for both tasks without invoking specifics of a particular task?
\begin{figure*}[!t]
    \centering
    \includegraphics[width=\textwidth]{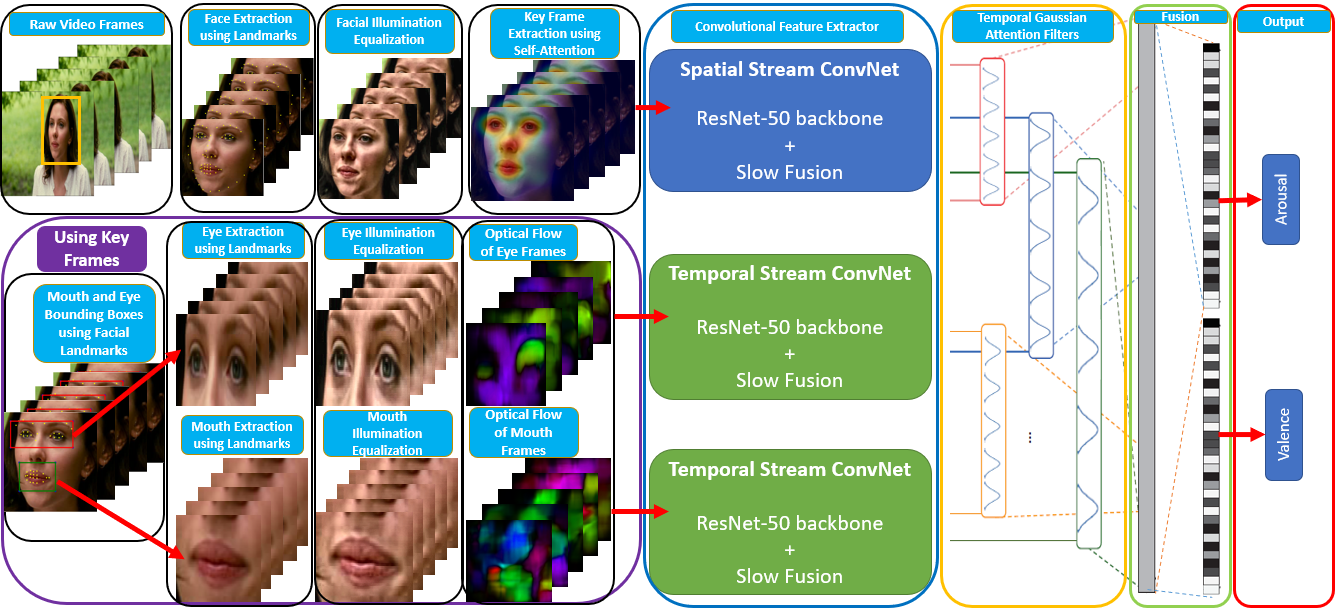}
    \caption{Pipeline of our Emotion Recognition System}
    \label{fig:pipeline}
\end{figure*}
\par In this work, we aim to explore the problem of emotion recognition from the perspective of action recognition. We use a continuous affect recognition dataset collected in-the-wild, called \textit{AFEW-VA} which consists of highly accurate per-frame annotations
of valence and arousal for 600 challenging, real-world video
clips extracted from feature films Added to these are per-frame annotations of 68 facial
landmarks. The dataset has been made publicly available.
\par We conduct a quantitative study of multiple standard and state-of-the-art emotion and action recognition algorithms applied to our dataset. We propose a novel three stream ensemble deep learning pipeline shown in Figure \ref{fig:pipeline}. The pipeline is designed as a regression system to detect continuous affect.Illumination equalization is performed across all video clips using Contrast Limited Adaptive Histogram Equalization (CLAHE). This helps a uniform illumination over all subjects, hence increasing the performance of the system.  Action recognition research \cite{wang2016temporal} shows that not all video frames are necessary for video classification.  Hence, a spatial self-attention mechanism is applied to compute the key frames which capture the same spatio-temporal information as that of the original video frames. The frame-wise Gaussian heatmaps provided by the attention network is used to identify the most discriminating regions of the face for efficient emotion recognition. Optical flow from eye and mouth regions, which get the highest score from attention maps, are extracted to capture temporal context in the dataset. Three streams of input, thus consists of RGB video frames, optical flows from eye and mouth.  A Convolutional feature extractor with ResNet-50 backbone is used to extract features from all the three streams. Further, variable length temporal Gaussian attention filters are used for capturing temporal attention. The output from the temporal attention filters are given to Fully Connected layers to get the final outputs.
\par The contributions of our work are as follows:
\begin{enumerate}
    \item A novel end-to-end three stream ensemble deep learning regression pipeline for continuous affect recognition.
    \item Key frame extraction and identification of the most discriminative features of the face using spatial self-attention.
    \item Using optical flow frames of eye and mouth regions to capture temporal context.
    \item Incorporation of Temporal Gaussian Attention filters into the three stream pipeline. 
    \item A quantitative comparative study of multiple standard and state-of-the-art action and emotion recognition algorithms applied to our chosen dataset .
\end{enumerate}

The rest of the paper is organized as follows: Section \ref{sec:related_work} mentions a brief literature survey regarding the evolution of emotion and action recognition algorithms and the work that has been implemented for this paper. Section \ref{sec:data} provides information about the chosen dataset. Section \ref{sec:approach} provides information about our pipeline. Section \ref{sec:results} provides information about the baseline algorithms that have been compared with our method and quantitative results of the comparative study. Section \ref{sec:conclusion} provides a summary and discussion of future work. Finally, Section \ref{sec:ack} provides acknowledgments for our work. 

%% file: related.tex
\section{Related Work}\label{sec:related_work}
There has been significant work on audio-based valence and arousal estimation \cite{4_gunes2013categorical}. Our paper focuses on video-based valence and arousal estimation. Both valence and arousal are defined
as continuous emotional dimensions. Therefore, it seems suitable to
study them directly in the continuous domain. Even though much
of the early work considered coarse levels of valence and arousal
(e.g. positive vs negative), and posed the problem as one of classification
\cite{10_corneanu2016survey}, more recent work casts the problem in the continuous
domain \cite{4_gunes2013categorical, 6_sariyanidi2014automatic, 11_mollahosseini2017affectnet}. 
\par With the onset of AVEC challenges in 2011, there has been tremendous progress in the field. Pioneering work was done with a subset of the Semaine dataset \cite{schuller2011avec}, which originally formulated the problem as one
of classification, using binarized values (±1), before moving to continuous
annotations in 2012. The
best results obtained that year were an average Pearson Correlation
Coefficient (PCC) of 0.456 \cite{nicolle2012robust}. The 2013 and 2014 editions of the
challenge used the audio-visual depression language corpus \cite{valstar2013avec, valstar2016avec}.
The best results obtained on that corpus were lower than in the
first year with a PCC of 0.1409 in 2013 \cite{meng2013depression} but improved to 0.5946
for the best performer in 2014 \cite{kachele2014inferring}. RECOLA dataset was used from 2015 in AVEC challenges. The results were good even though RECOLA \cite{ringeval2015av+} is a challenging dataset. For instance, the best performer
in the 2015 edition of the challenge obtained an average PCC of
0.685 \cite{he2015multimodal}, while the best performer in 2016 obtained an average
PCC of 0.731 \cite{brady2016multi}.
\par The most obvious approaches, which can also be treated as non-deep learning baselines to valence and arousal estimation are static regression, such as linear regression \cite{gupta2014multimodal, tzirakis2017end}, partial least squares \cite{meng2013depression}, and Support Vector Machine for Regression (SVR) \cite{nicolaou2011continuous, sanchez2013audiovisual, kossaifi2017afew}. In \cite{sanchez2013audiovisual}, SVR is combined with Canonical Correlation Analysis (CCA) to iteratively fuse predictions. In \cite{kossaifi2017afew}, SVR is used as a baseline to test the performance of multiple facial appearance and geometric features. SVR is also employed in the work of \cite{kachele2014inferring} but its use differs in that
template trajectories for each emotional dimension are first built and
then matched to a new testing sample using metadata as features.
Meta-data is also used in \cite{soladie2012multimodal} where audio, video and contextual
(meta-data) features are combined in a multimodal fuzzy inference
system. More powerful kernel based methods can be used such as
Nadaraya-Watson kernel regression \cite{nicolle2012robust} or Doubly Sparse Relevance
Vector Machine \cite{kaltwang2015doubly} that can impose sparsity on both the kernels and
the training samples.
\par In \cite{kossaifi2017afew}, a comparative study has been done on multiple appearance and geometric features. In appearance features, the paper experiments with Hybrid-SIFT \cite{al2016facial}, Block LBP \cite{zhang2007face}, Hybrid LBP, Holistic-DCT \cite{mohanraj2016robust}, Block-DCT and Hybrid-DCT. Normalized features \cite{kim2001real} are explored as geometric features. When dealing with several features, e.g. geometric and appearance
features, or multiple modalities, e.g. audio and video, the question
of how to fuse them arises. There are four main types of fusions:
feature level (early fusion), decision level (late fusion), model-level
and output-associative fusion \cite{zeng2008survey, pantic2003toward}. In the early-fusion case, the
features are combined most commonly by simply concatenating them
and using the output for estimation(e.g.as is done in \cite{chen2015multi} for predicting
valence). However, in general, this approach tends to create very high
dimensional feature vectors and lead to overfitting.
Late fusion is the process of first generating separate estimations
from each modality before combining – fusing – them into one final
estimation. This fusion can be achieved in numerous ways, from simple
mapping, e.g. averaging, as in \cite{kachele2015ensemble}, to more complex methods
such as linear and multi-linear regression [15,25,26,28,30,37,38],
SVR \cite{kossaifi2017afew}, random forests \cite{cardinal2015ets} or Kalman filters based \cite{brady2016multi}. This
type of late fusion is akin to stacking, a classical ensemble learning
technique.
\begin{figure}[h]
    \centering
    \includegraphics[width=\linewidth]{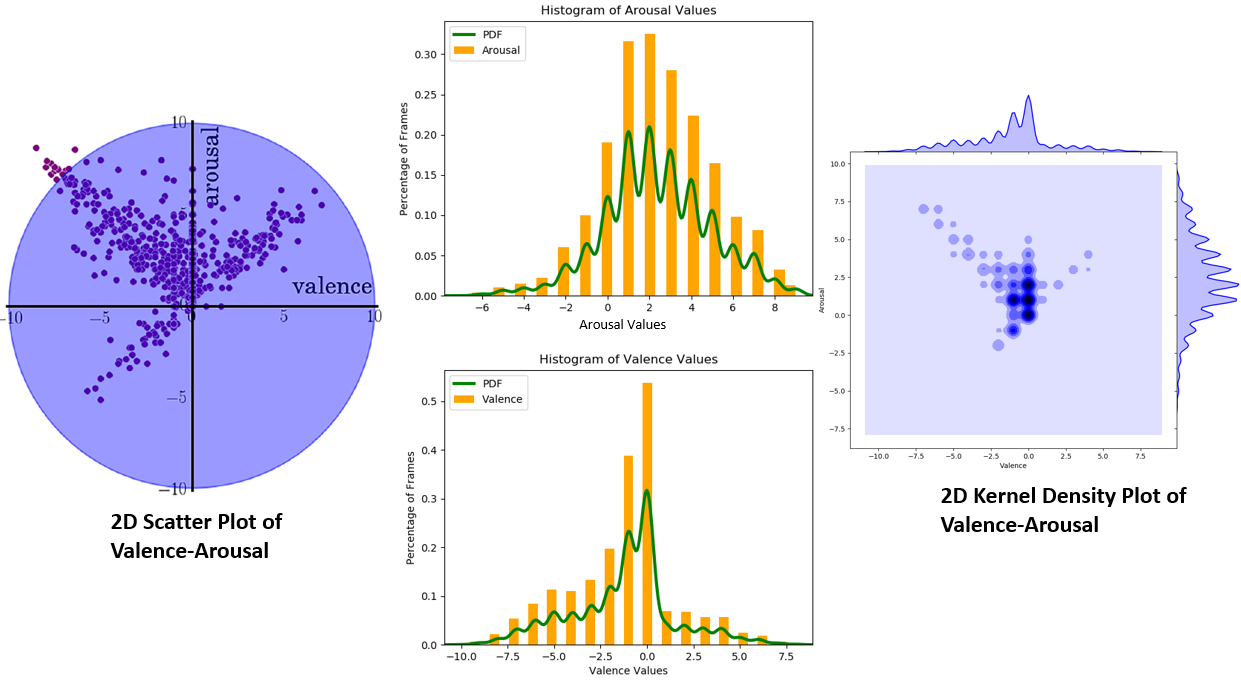}
    \caption{Data Statistics of our dataset}
    \label{fig:data_stat}
\end{figure}

\par Deep learning approaches to solve the problem of continuous affect use blocks of convolutional neural networks as feature extractors with SVM/SVR as classifiers, or have an end-to-end strategy of training. A combination of convolutional and recurrent neural networks is also used to capture spatio-temporal context. The most commonly used metric is CCC (Concordance Correlation Coefficient) which is a metric to penalize the shift in time series predictions. Mean Squared Error is also a commonly used metric to evaluate the performance of continuous affect models. \cite{benroumpi2019affwild} uses a network called AffWildNet whihc consists of convolutional and pooling layers of either VGG-Face or ResNet-50 structures, followed by a fully connected layer and two RNN layers with GRU units. This is used as the deep learning baseline for our paper, as this paper was one of the most recent studies of continuous affect using deep learning \cite{nagendra2017comparison, nagendra2022constructing, funk2018learning,liu2021new,pei2021utilizing,nagendra2020cloud,nagendra2020efficient,nagendra2022threshnet,nagendra2024patchrefinenet,nagendra2023estimating,zhu2022rapid}, which was published in 2019. 
\par Before deep learning came along, most of the traditional Computer Vision algorithm variants for action recognition can be broken down into the following 3 broad steps:
(1) Local high-dimensional visual features that describe a region of the video are extracted either densely or at a sparse set of interest points. (2) The extracted features get combined into a fixed-sized video level description. One popular variant to the step is to bag of visual words (derived using hierarchical or k-means clustering) for encoding features at video-level. (3) A classifier, like SVM or RF, is trained on bag of visual words for final prediction.
\par Of these algorithms that use shallow hand-crafted features in (1), improved Dense Trajectories \cite{wang2013action} (iDT) which uses densely sampled trajectory features was the state-of-the-art. Simultaneously, 3D convolutions were used as is for action recognition without much help in 2013 \cite{ji20123d}. Soon after this in 2014, two breakthrough research papers were released which form the backbone for all the papers we are going to discuss in this post. The major differences between them was the design choice around combining spatio-temporal information.
\par This \cite{karpathy2014large} is on of the foremost attempts of video classification using deep learning. The paper explores multiple ways to fuse temporal information from consecutive frames using 2D pre-trained convolutions. the consecutive frames of the video are presented as input in all setups. Single frame uses single architecture that fuses information from all frames at the last stage. Late fusion uses two nets with shared parameters, spaced 15 frames apart, and also combines predictions at the end. Early fusion combines in the first layer by convolving over 10 frames. Slow fusion involves fusing at multiple stages, a balance between early and late fusion. For final predictions, multiple clips were sampled from entire video and prediction scores from them were averaged for final prediction. The slow fusion model, which performed the best according to the authors of this paper, is used as the first action recognition baseline in our work. In this work \cite{simonyan2014two}, the authors build on the failures of the previous work by Karpathy et al. Given the toughness of deep architectures to learn motion features, authors explicitly modeled motion features in the form of stacked optical flow vectors. So instead of single network for spatial context, this architecture has two separate networks - one for spatial context (pre-trained), one for motion context. The input to the spatial net is a single frame of the video. Authors experimented with the input to the temporal net and found bi-directional optical flow stacked across 10 successive frames was performing best. The two streams were trained separately and combined using SVM. The final prediction was the same as the previous paper, i.e. averaging across sampled frames. This work is used as the second action recognition baseline in our paper. In this work \cite{Feichtenhofer_2016},  authors use the base two-stream architecture with two novel approaches and demonstrate performance increment without any significant increase in the size of parameters. The authors explore the efficacy of two major ideas: (1) Fusion of spatial and temporal streams (how and when) - For a task discriminating between brushing hair and brushing teeth - a spatial net can capture the spatial dependency in a video (if it’s hair or teeth) while temporal net can capture the presence of periodic motion for each spatial location in the video. Hence it’s important to map spatial feature maps about say a particular facial region to a temporal feature map for the corresponding region. To achieve the same, the nets need to be fused at an early level such that responses at the same pixel position are put in correspondence rather than fusing at the end (like in base two-stream architecture). (2) Combining temporal net output across time frames so that long-term dependency is also modeled.  Our architecture is designed using this framework. Finally, \cite{Wang_2016} provides strategies for good practices in action recognition: (1) They suggest sampling clips sparsely across the video to better model long-range temporal signal instead of random sampling across the entire video. (2) For the final prediction at the video-level authors explored multiple strategies. The best strategy was: (a) Combining scores of temporal and spatial streams (and other streams if other input modalities are involved) separately by averaging across snippets (b) Fusing scores of final spatial and temporal scores using weighted average and applying softmax over all classes. These points have been incorporated while designing our pipeline.

%% file: data.tex
\section{AFEW-VA: emotion recognition dataset in-the-wild}\label{sec:data}
This section provides information about the dataset that has been used for all the experiments in this paper.
\subsection{Data}
\begin{figure}
    \centering
    \includegraphics[width=\linewidth]{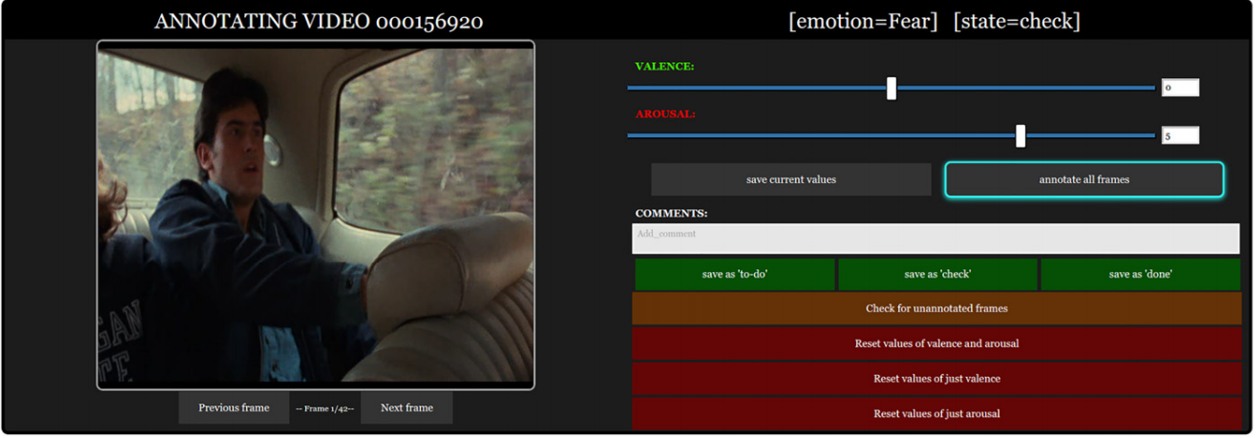}
    \caption{Screenshot of the annotation tool developed and used to annotate the AFEW-VA dataset.}
    \label{fig:ann}
\end{figure}
\begin{figure}
    \centering
    \includegraphics[width=\linewidth]{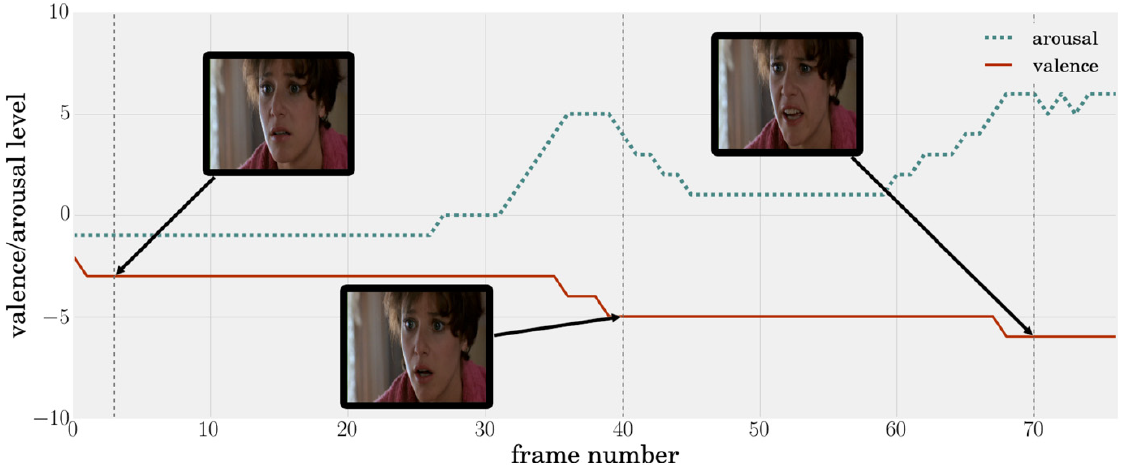}
    \caption{Example of annotated valence and arousal levels for a sample video from our dataset along with some representative frames.}
    \label{fig:temp_eval}
\end{figure}
\begin{figure}
    \centering
    \includegraphics[width=\linewidth]{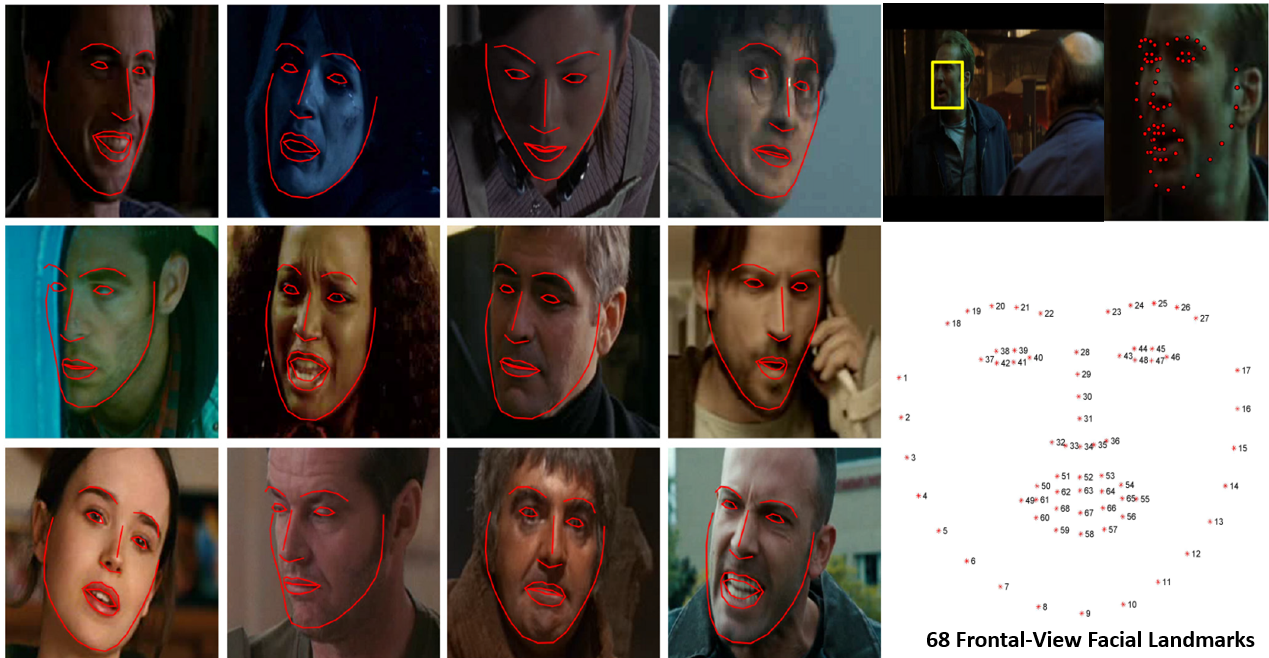}
    \caption{Examples of tracked landmarks in the dataset}
    \label{fig:landmarks}
\end{figure}
AFEW-VA consists of 600 videos extracted from features films.
The videos are processed at 30fps and range from
short (around 10 frames) to longer clips (more than 120 frames),
and display various facial expressions. They are captured under challenging
indoor and outdoor conditions such as complex cluttered backgrounds, poor illumination, large out-of-plane head rotations, variations in scale, and occlusions. In total, there are $30000$ frames with per-frame levels of valence and arousal intensities in the range of $−10$ to $10$. Figure \ref{fig:data_stat} shows the distribution of the values of valence and arousal present in the dataset. It matches the expected distribution and, as can be
seen, there is a wide range of values for both valence and arousal. In
some videos, we observe a significant signal change in valence and
arousal across the frames. In some other videos, the temporal change
of valence and arousal is negligible. Figure \ref{fig:data_stat} also shows the distribution of
the annotations in the valence and arousal circle. As can be observed,
our data show large variations in valence and arousal values and
complement well existing databases.

\subsection{Annotations}
Per-frame annotations of valence and arousal levels are provided for all frames within all video clips of our dataset. Annotations are created by two expert annotators, a male and a female, FACS AU coding certified. Both annotators have annotated all videos together, therefore discussing all disagreements and
coming up with a unique solution. Hence, AFEW-VA annotations are
detailed and highly accurate. The range of annotation levels for both
valence and arousal is from −10 to 10, resulting in a total of 21
levels. Figure \ref{fig:ann} shows a screenshot of the annotation tool used by the experts. Figure \ref{fig:temp_eval} shows the temporal evolution of the valence and arousal
signals for an example video from our dataset, along with some representative
frames. The dataset also provides accurate location for 68 landmarks, as shown in Figure \ref{fig:landmarks}, including both interior and boundary points of the face. The facial annotations were generated in a semi-automatic way. First, the face was detected in the first frame of each video sequence using the
tree-based deformable part model (DPM).

%% file: approach.tex
\section{Our Approach}\label{sec:approach}
We propose a novel three stream ensemble deep learning pipeline for emotion recognition using concepts of action recognition mechanisms. In this section, the proposed pipeline shown in Figure \ref{fig:pipeline} will be discussed. 
\subsection{Key Frame Sub-sampling}
\begin{figure}
    \centering
    \includegraphics[width=3.5in]{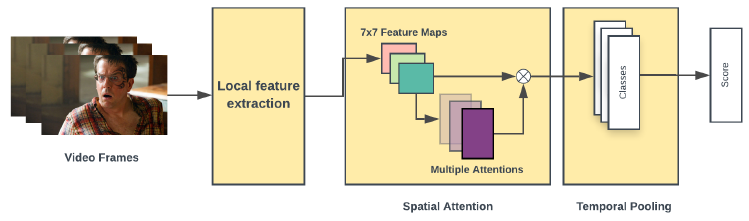}
    \caption{The CNN takes the video frame as its input and produces local features. Using
the local features, the multi-head attention network computes the weight importance
of each local image feature. The aggregated representation is computed by multiplying
multi-head attention output and the local image features. This representation is then propagated through temporal softmax pooling to extract global features over the entire
video.}
    \label{fig:sp_attn}
\end{figure}

\begin{figure*}
    \centering
    \includegraphics[width=\textwidth]{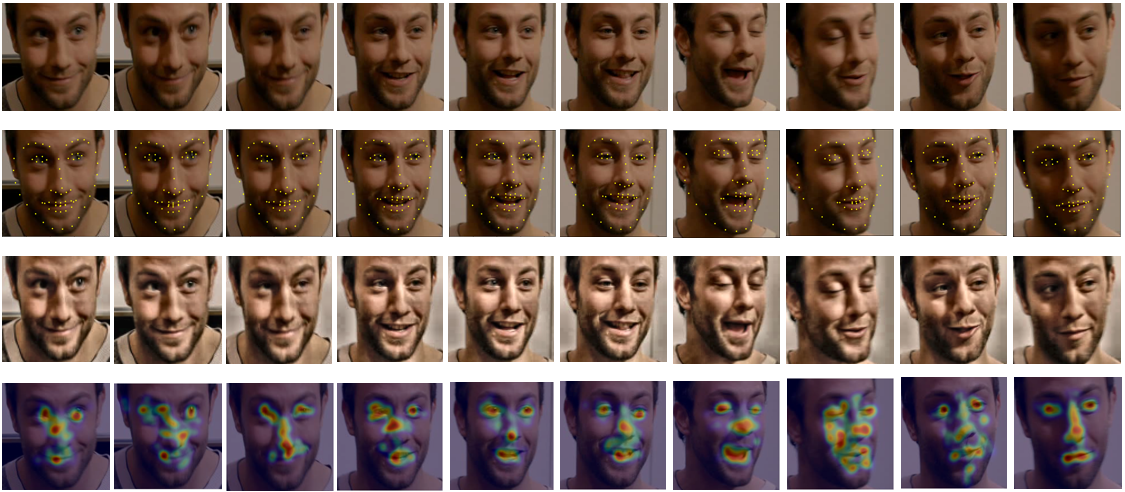}
    \caption{Steps of Key Frame subsampling. Row 1: Raw face extracted images. Row2: Facial Landmarks for faces. Row3: Illumination Equalization using CLAHE Row4: Sub-sampled Key-Frames and their 2D Gaussian attention heatmaps showing most important facial regions for efficient emotion recognition.}
    \label{fig:face_attn}
\end{figure*}
Each video clip has variable number of video frames, which makes it difficult to design a deep learning architecture. As described in Section \ref{sec:related_work}, \cite{Wang_2016} tells us that we do not need all the frames of every video clip to get a high performing architecture.  Instead of randomly sub-sampling video frames, we incorporate a spatial self-attention mechanism to sample equal number of key discriminative frames from each video clip. Every video clip, as shown in Figure \ref{fig:landmarks} shows varying illumination and distance from camera perspective of subjects in every video clip. This poses a problem to capture spatio-temporal context for facial expression recognition efficiently. The reasons for not exploring body language are:
    (1) There was no publicly available in-the-wild continuous affect dataset to perform experiments. (2) There were not enough deep learning baselines for a comparative study. (3) Emotions are highly correlated with the face.
\par Using the 68 facial landmarks that were provided, faces are extracted from every video clip by drawing a bounding box covering all the landmarks. Illumination of every video frame is equalized by using Contrast Limited Adaptive Histogram Equalization (CLAHE) \cite{zuiderveld1994contrast}. Adaptive histogram equalization (AHE) is a computer image processing technique used to improve contrast in images. It differs from ordinary histogram equalization in the respect that the adaptive method computes several histograms, each corresponding to a distinct section of the image, and uses them to redistribute the lightness values of the image. It is therefore suitable for improving the local contrast and enhancing the definitions of edges in each region of an image. However, AHE has a tendency to over-amplify noise in relatively homogeneous regions of an image. A variant of adaptive histogram equalization called contrast limited adaptive histogram equalization (CLAHE) prevents this by limiting the amplification. 

\subsubsection{Local Feature Extraction}
\par The idea of key frame sub-sampling is taken from the paper \cite{Aminbeidokhti_2019}. We have frame-by-frame annotation of our dataset. This enabled us to apply this approach to capture the most important frames for regression. Consider a video sample $\mathbf{S_i}$ and its associated emotion $y_i \in R^E$, where we represent the video as a sequence of $F$ frames $[X_{0,i}, ...X_{F,i}]$, each of size $W\times H \times3$. VGG-16, pre-trained with VGG-Face model \cite{parkhi2015deep} was used for extracting an independent description of a face on each frame in the video. For a given frame $X$ of a video, we consider the feature map produced by the last
convolutional layer of the network as representation. This feature map has spatial
resolution of $L=H/16 \times W/16$ and $D$ channels. channels.We discard the spatial resolution
and reshape the feature map as a matrix $R$ composed of $L$ $D$-dimensional local
descriptors (row vectors).
\begin{equation}
    \mathbf{R} = VGG_{16}(X)
\end{equation}
These descriptors will be associated to a corresponding weight and used for the attention mechanism.
\subsubsection{Spatial Attention}
For the spatial attention we rely on the self-attention mechanism \cite{vaswani2017attention}, which
aggregates a set of local frame descriptors $R$ into a single weighted sum $v$ that
summarizes the most important regions of a given video frame: 
\begin{equation}
    v = a\mathbf{R}
\end{equation}
where $a$ is a row vector of dimension $L$, which defines the importance of each frame region. The weights $a$ are generated by two layers fully connected network that associates each local feature (row of $R$) to a corresponding weight. This vector representation usually focuses on a specific region in the facial
feature, like the mouth. However, it is possible that multiple regions of the face
contain different type of information that can be combined to obtain a better
idea of the person emotional state. Hence, we need multiple attention units that focus on
different parts of the image. For doing that, we transform $w_{s2}$ into a matrix
$W_{s2}$ of size $R \times L$, in which every row represents a different attention.
\subsubsection{Temporal Pooling}
After extracting the local features and aggregating them using the attention
mechanism for each frame, we have to take into account frame features over the whole video. As the length of a video can be different for each
example, we need an approach that supports different input lengths. The most
commonly used approaches are average and max pooling; however, these techniques assume that every frame of the video has the same importance in the final decision (average pooling) or that only a single frame is considered as a
general representation of the video (max pooling). In order to use the best of
both techniques, we use an aggregation based on softmax, which can be considered a generalization of the average and max pooling.
\par Given a video sample $\textbf{S}$, after feature extraction and spatial attention, we obtain a matrix $V$ in which each row represents the features of a frame. The 21 labels of valence and arousal are discretized into 7 classes based on the 2D continuous emotion wheel for this process. These features are converted into class scores
through a final fully connected layer $\mathbf{O} = \mathbf{W_{sm}V}$. Here, $\mathbf{O}$ is a matrix in which an element $o_{c,f}$ is the score for class $c$ of the frame $f$. We then
transform the scores over frames and classes in probabilities with a softmax.

\begin{equation}
    p(c, f|S) = \frac{exp(o_{c,f})}{\sum_{j,k}exp(o_{j,k})}
\end{equation}
In this way, we obtain a joint probability on class $c$ and frame $f$. This representation can be marginalized over classes $p(f|S) = \sum_{c}p(c, f|S)$. This will give us information about the most
important frames of a given video. The activations of each frame are converted to 2D Gaussian heatmaps, which will give us a qualitative evaluation of the most important regions of the face as shown in Figure \ref{fig:face_attn}. The figure shows all the steps of key frame sub-sampling. Using this method, we sample 10 key frames for each video clip. 10 key frames with highest classification scores are chosen.

\subsection{Optical Flow}
\begin{figure*}
    \centering
    \includegraphics[width=\textwidth]{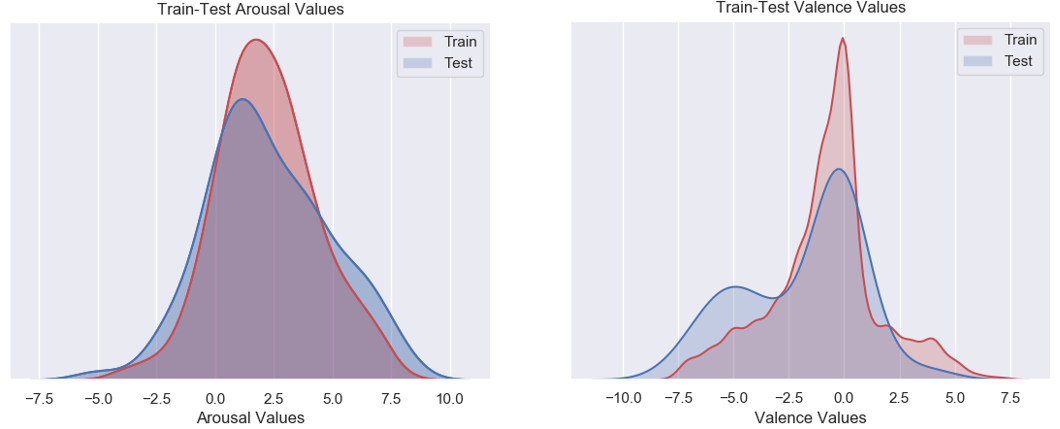}
    \caption{Train-Test Distributions of Arousal and Valence}
    \label{fig:train_test}
\end{figure*}
Key Frame sub-sampling shows us that eyes and mouth are the most important regions of the face, from which discriminative features can be extracted. Hence, as shown in Figure \ref{fig:pipeline}, we use the extracted key frames and facial landmarks to automatically draw bounding boxes over the mouth and eyes of every frame. This way, the mouth and eyes are extracted from every key frame. Now, we have RGB images video frames and RGB frames of mouth and eyes. 
\par Optical flow \cite{horn1981determining} is the pattern of apparent motion of image objects between two consecutive frames caused by the movemement of object or camera. It is 2D vector field where each vector is a displacement vector showing the movement of points from first frame to second. Optical flow is explicitly used in Action Recognition algorithms to extract motion information. To incorporate action recognition mechanisms and to effectively capture temporal context, we extract optical frame information between every two consecutive frames. After this step, we have our input data for any action recognition mechanism that is rich in spatial and temporal context. 

\subsection{Temporal Gaussian Attention Filters}

\begin{figure}
    \centering
    \includegraphics[width=3.5in]{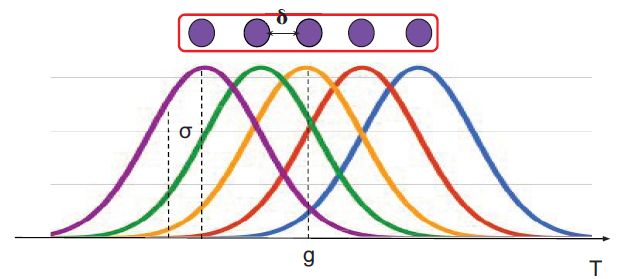}
    \caption{An illustration of our temporal attention filter. The
filter is differentiable and represented with three parameters.}
    \label{fig:temp_attn}
\end{figure}

Many high-level activities are
often composed of multiple temporal parts (e.g., sub-events)
with different duration/speed, and our objective is to make the
model explicitly learn such temporal structure using multiple
attention filters and benefit from them. The idea of temporal filters is taken from \cite{piergiovanni2017learning}. The filters are designed to be fully differentiable, allowing end-to-end training. We can learn a set of optimal static temporal attention filters to be shared across different videos. Hence, in our work, we incorporate this idea to capture global temporal attention between video clips.
\par The temporal attention filters are applied after we do convolutional feature extraction from segments of each video clip, as shown in Figure \ref{fig:pipeline}. Each temporal filter learns three parameters: a center $g$, a stride $\delta$ and a width $\sigma$. These parameters determine where the filter is placed and the size of the interval focused on. A filter consists of $N$ Gaussian filters separated by a astride of $\delta$ frames. The goal of this model is to learn where in the video the most useful features appear. 
\begin{figure*}[h]
    \centering
    \includegraphics[width = \textwidth]{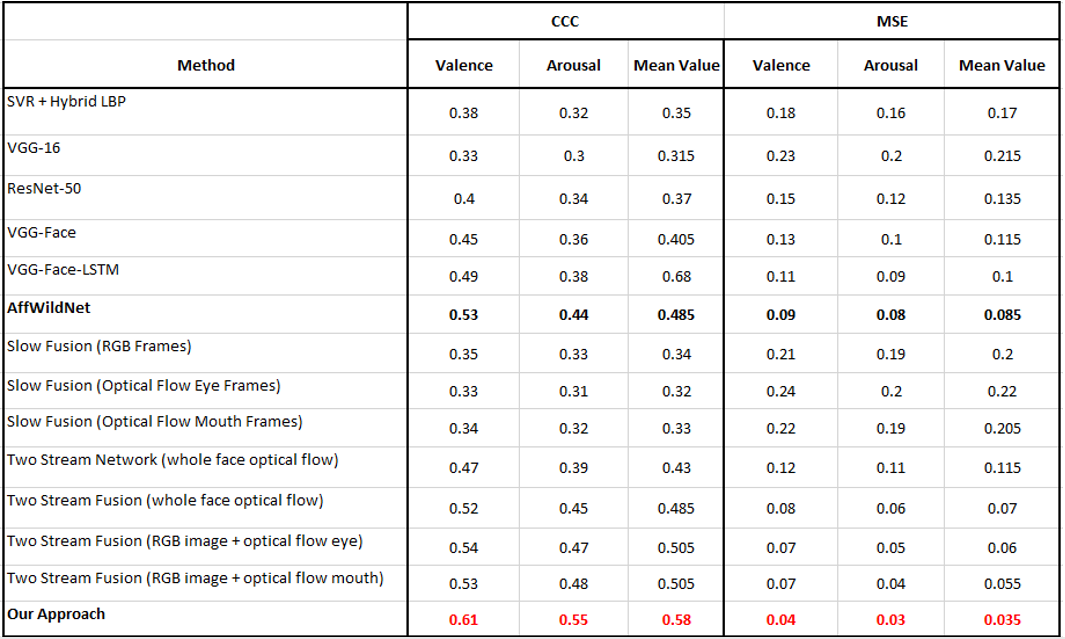}
    \caption{Table of quantitative results}
    \label{fig:results}
\end{figure*}

\subsection{Network Architecture and Training}

We propose a three stream convolutional feed-forward architecture for feature extraction. The inputs to the network are key facial RGB frames, optical flow frames of eyes and optical flow frames of mouth. All the inputs are of shape 96 X 96 X 3. Every block of the architecture consists of {Conv - Batch Normalization - Max Pooling} layers. The network has one spatial stream which takes RGB images as input and two temporal streams which take optical flow as inputs. The architecture is an encoder with decreasing resolution and increasing number of channels with depth. Dropout regularization is used to avoid over-fitting. Out of a total of 600 video clips, 500 are used for training and 100 are for test. The train and test video clips are chosen such that they have identical independent frame-wise valence-arousal distributions as shown in Figure \ref{fig:train_test}. The features from the three streams are fused using average pooling after after 5 blocks. The fused features are given to temporal attention filters, after which there are two fully connected blocks. 21 levels of Arousal and Valence [-10, 10] are converted to values between {0,1} by using min-max normalization. All layers use ReLU activation, while the last layer uses sigmoid activation. 1 - (Average of Concordance Correlation Coefficient (CCC) or Arousal and Valence) is used as the Loss function. The network is trained using Nvidia RTX 2080 Ti GPU for 200 epochs with a batch size of 32. We use a Learning Rate Scheduler with an initial learning rate of 5e-5.     

%% file: results.tex
\section{Results}\label{sec:results}
In this section we are going to discuss quantitative results and the baselines used. Table \ref{fig:results} shows the quantitative scores of CCC and MSE. It can be observed that Our Approach performs the best on afew-va dataset over all other emotion and action recognition models. This is because our approach is an ensemble approach that includes sub-modules of multiple action recognition algorithms, such as optical flow streams, parallel feature extraction and temporal attention filters. On top of this, key frame extraction using spatial self-attention ensures the accuracy with discriminative frames with a low processing overhead.

\subsection{Baselines}
We use Support Vector Machine for Regression (SVR) as our baseline. Hybrid LBP features are extracted from Facial video frames. LBP is a visual descriptor used for texture classification. We use a simple linear transformation model to scale and translate all the images onto a common coordinate frame. We extract 27X27 patches around each of the 68 landmarks. Each patch is represented by a 59-D histogram. All histograms are concatenated to form a global descriptor of size 4012. SVR is trained on these features to establish a baseline. The CNN-RNN model called AffWildNet, mentioned in Section \ref{sec:related_work} is used as our deep learning baseline for the task of emotion recognition. Further, all the action recognition papers mentioned in Section \ref{sec:related_work} are implemented on our dataset for comparative analysis. 

\subsection{Evaluation Metrics}
Two evaluation metrics are used in this paper for quantitative comparative analysis:
\subsubsection{Concordance Correlation Coefficient (CCC)}
\begin{equation}
    \rho_c = \frac{2s_{xy}}{s_{x}^{2} + s_{y}^{2} + (\mu_x - \mu_y)^2}
\end{equation}, where 
$s_x$ and $s_y$ are standard deviations and $\mu_x$ and $\mu_y$ are corresponding means.
This is a widely used metric to measure the performance of dimensional emotion recognition methods.  CCC evaluates the agreement between two time-series data, in our case, video annotations and predictions. CCC is calculated by scaling the mean square difference between ground truth and predictions. Predictions that are well correlated with annotations, but shifted in value are penalized in proportion to the deviation. CCC takes the values in the range [-1,1]. +1 indicates perfect concordance, while -1 indicates perfect discordance. 
The loss function is given by
\begin{equation}
    Loss = 1 - \frac{\rho_a + \rho_v}{2}
\end{equation}

\subsubsection{Mean Squared Error}
\begin{equation}
    MSE = \frac{1}{N}\sum_{i=1}^{N}(x_i - y_i)^2
\end{equation}
This is a common evaluation metric used for regression problems. 

%% file: conclusion.tex
\section{Conclusion and Future Work}\label{sec:conclusion}
This paper presents a novel three-stream action recognition-based emotion recognition pipeline with spatial key frame extraction mechanism and temporal Gaussian filters. This is an ensemble approach that combines multiple sub-modules of standard and state-of-the-art action recognition algorithms to get state-of-the-art accuracy on the continuous, in-the-wild affect dataset AFEW-VA. Further, our work is a proof of concept that the action recognition mechanisms can be extrapolated to the task of continuous emotion recognition. To the best of our knowledge, this is the first multi-stream pipeline with spatial self-attention, optical flow inputs of facial regions and temporal Gaussian Attention filters to be designed to solve the problem of continuous affect recognition. Further, the paper provides a detailed comparative study to compare multiple standard and state-of-the-art emotion and action recognition algorithms. 
\par For future work, the whole body could be considered for expression recognition by extracting bodily features. This would be a more challenging problem. The model could be fine-tuned further for better accuracy. Adversarial training and Semi-supervised training can be used for further improvement in accuracy.

%% file: ack.tex
\section{Acknowledgements}\label{sec:ack}
We would like to thank Dr. James Wang for all his support. The course was a great learning opportunity and a stepping stone to the interesting domain of emotion recognition. We would like to thank I-Bug for the dataset.